# I2R-MI at #SMM4H 2023: A Generic NLI approach for Classification of Sentiment Associated with Therapies

Rajaraman Kanagasabai[1] and Anitha Veeramani[1]

[1]Institute for Infocomm Research, Agency for Science Technology and Research, Singapore
Email: {kanagasa, vanitha}@i2r.a-star.edu.sg

**Abstract**

*This paper describes our system for addressing SMM4H 2023 Shared Task 2 on 'Classification of sentiment associated with therapies (aspect-oriented)'. In our work, we adopt an approach based on Natural language inference (NLI) to formulate this task as a sentence pair classification problem, and train transformer models to predict sentiment associated with a therapy on a given text. Our best model achieved 75.22% F1-score which was 11% (4%) more than the mean (median) score of all teams' submissions.*

**Introduction**

There is an abundance of health-related data on social networks, including chatter about therapies for health conditions. These therapies include but are not limited to medication, behavioral, and physical therapies. Social media subscribers who discuss such therapies often express their sentiments associated with the therapies. SMM4H 2023 [1] Task 2 aims to capture this opportunity by drawing tweet posts from multiple pre-identified Twitter cohorts (chronic pain, substance use disorder, migraine, chronic stress, long-COVID, and intimate partner violence), and enable development of systems that can classify sentiments associated with a therapy into one of three classes—*positive, negative, and neutral*. It may be noted that a typical sentiment classification approach is inadequate, since one needs consider the 'therapy' along with the tweet to predict the sentiment label. In our work, we adopt an approach based on Natural language inference (NLI) to address this *aspect*-oriented sentiment classification problem task.

Natural language inference involves detecting inferential relationships between a 'premise' text and a 'hypothesis' text, and is considered fundamental in natural language understanding (NLU) research [2,3]. The objective in NLI is to determine whether hypothesis $h$ is true ('entailment'), false ('contradiction'), or un-determined ('neutral') given the premise P. Our main idea is to formulate the sentiment classification problem as an NLI task with sentiment classes as inference labels.

**Methods**

The challenge training data consists of 5000 English Tweets containing mentions of a variety of therapies manually labeled as positive, negative, or neutral with the following approximate distribution: 20%, 14%, and 66%, respectively.

We first preprocessed the tweets by converting user mentions and web/url links into @USER and HTTPURL, respectively. We also translated emotion icons into text strings.

Given a tweet and associated therapy, we treat the tweet text as 'premise' and a statement on therapy as 'hypothesis', to formulate as an NLI task. For illustration, 'This is on {therapy}' was our hypothesis template.

We randomly split the instances into 80% training and 20% dev for model development. We implemented fine-tuned BERT [4] and RoBERTa [5] base models and used '*[CLS] tweet text [SEP] hypothesis [SEP]*' as processed input to the transformers to predict the sentiment classes. To handle class imbalance, we under-sampled from the largest class ('neutral') and we chose 1:1:2 class ratio after experimentation.

The experiments were executed on NVIDIA-GeForce Tesla V100 series SXM2-32GB with 5 cores of GPU machines. Models were trained over 10 epochs for training and validation. The pretrained weights for the transformers prior to fine-tuning were from the HuggingFace NLP Library.

**Results**

We conducted several experiments and the final configuration chosen was as follows. BERT model uses 12-layer, 768-hidden, 12-heads, 110M parameters for base version and RoBERTa uses 123M parameters. The models are trained end-to-end using AdamW optimizer with the decay rate of 0.9. In addition, we have experimented with different learning rates to understand if there is any change in performance. However, learning rate of 5e-6 shows a steady linear increase with the specified decay rate for RoBERTa model.

The best score on dev set was ~76% with BERT and ~79% with RoBERTa, where the random sampling baseline score was 66%. Thus, we used the two models to generate test set submissions. The final results are presented in Table 1.

Table 1. Summary of Performance on Test Data

| Models | Test F1-Score (%) |
|---|---|
| BERT-base NLI | **75.2%** |
| RoBERTa-base NLI | 75.1% |
| Median Score (ALL TEAMS) | 71.0% |
| Mean Score (ALL TEAMS) | 64.0% |

We observe that, with a simple NLI approach, we achieved 75.2% micro-averaged F1-score which was 11% (4%) more than the mean (median) score of all teams' submissions. Furthermore, we just used 'base' models and no external data or domain knowledge, and thus our methodology is promising as a generic approach for aspect-oriented sentiment classification.

**Experiments on SMM4H 2023 Shared Task 5**

SMM4H 2023 Shared Task 5 aims to capture the potential of early detection and novel discovery of Adverse Drug Events (ADE's), and addresses the task of extraction and normalization of ADEs in tweets to their standard concept IDs in the Medical Dictionary for Regulatory Activities (MedDRA) vocabulary. The challenge training data consists of 18,000 labeled tweets, along with the tweet text, the annotated binary label for whether or not the tweets contains an ADE, the character offsets of the ADE, the text span of the ADE, and the MedDRA ID.

For our experiments, we preprocessed the tweets as in Task 2, and trained a seq2seq model using a T5 transformer [6] to extract the ADE text spans from the tweet. Then, for every tweet with ADE text span identified, we used SBERT model [7] to retrieve MedDRA entries for a given ADE text span based on cosine similarity measure. The top ranked MedDRA entry for every ADE text span is returned (if the similarity score is > 0.5) as the normalized ADE.

The results are presented in Table 2.

Table 2. Task 5 Performance on Test Data

| | Overall | | | Unseen | | |
|---|---|---|---|---|---|---|
| | F1 | Precision | Recall | F1 | Precision | Recall |
| Our Submission | 0.322 | 0.249 | 0.455 | 0.195 | 0.128 | 0.406 |
| Mean | 0.329 | 0.293 | 0.422 | 0.202 | 0.151 | 0.360 |
| Median | 0.322 | 0.249 | 0.405 | 0.195 | 0.128 | 0.354 |

Our method achieved F1 scores (Overall and Unseen) on par with the median scores of all teams' submissions. In particular, the recall was better than both mean and median scores, most notably under 'Unseen' ADE's but the precision scores were lower. This is possibly due to the naïve SBERT matching. It could be improved by fine-tuning on the training data, which will be explored in our future work.

**Discussion and Conclusions**

In this paper, we described our system for addressing 'SMM4H 2023 Shared Task 2' on 'Classification of sentiment associated with therapies (aspect-oriented)'. We adopted an NLI-based approach to address this task as a sentence pair classification problem, and fine-tuned transformer models to predict sentiment associated with a therapy on a given text. Our BERT model achieved 75.22% F1-score which was 11% (4%) more than the mean (median) score of all teams' submissions. We conclude that our simple approach using just 'base' models and no external data or domain knowledge, is promising for aspect-oriented sentiment classification. Possible scope for improvement include fine-tuning with more recent LLM's and pre-training on large tweet collections.

We also considered 'SMM4H 2023 Shared Task 5' on 'Normalization of adverse drug events in English tweets' and conducted experiments using a seq2seq model followed by SBERT based entity linking. Our approach resulted in higher recall even under zero-shot learning setup, though at a below-par precision. The results show promise but could be improved with further fine-tuning or advanced entity linking methodologies.